\pdfoutput=1
\documentclass[11pt, british]{article}
\usepackage[final]{acl}

\usepackage{times}
\usepackage{latexsym}
\usepackage[tone]{tipa}
\usepackage{tikz}
\usepackage{makecell}
\usepackage{rotating}
\usepackage{amssymb}
\usepackage{pifont} 
\usepackage{tabularray} 
\usepackage{graphicx}
\usepackage{array}
\usepackage{fdsymbol}
\usepackage{cleveref}
\usepackage{booktabs}
\usepackage{amsmath}
\usepackage{multirow}
\usepackage{xspace}
\usepackage{subcaption}
\usepackage{listings}

\usepackage{tablefootnote}
\usepackage{threeparttable}
\usepackage{rotating}

\usepackage{tcolorbox}
\usepackage{xcolor}

\usepackage[T1]{fontenc}

\usepackage[utf8]{inputenc}

\usepackage{microtype}
\usepackage{inconsolata}

\usepackage{twemojis}

\usepackage{tcolorbox}

\usepackage{tabularray}
\usepackage{cellspace}
\newcommand\cincludegraphics[2][]{\raisebox{-0.3\height}{\includegraphics[#1]{#2}}}
\usepackage{setspace}

\newcommand{\gpp}[0]{\texttt{G2P+}\xspace}
\newcommand{\ttipa}[1]{\texttt{\textipa{#1}}}
\newcommand{\phoible}[0]{Phoible\xspace}
\newcommand{\ipachildes}[0]{\textsc{IPA CHILDES}\xspace}

\title{\ipachildes \& G2P+: Feature-Rich Resources for \\ Cross-Lingual Phonology and Phonemic Language Modeling}

\author{
    {\bf Z\'{e}bulon Goriely} \texttwemoji{orange} ~~~~~ 
    {\bf Paula Buttery} \texttwemoji{orange}\texttwemoji{lemon} \\
    \texttwemoji{orange} Department of Computer Science \& Technology, University of Cambridge, U.K. \\
    \texttwemoji{lemon} ALTA Institute, University of Cambridge, U.K. \\
    \texttwemoji{orange} \texttt{firstname.secondname@cl.cam.ac.uk} \hspace{2mm}
}

\begin{document}
\maketitle
\begin{abstract}
In this paper, we introduce two resources: (i) \gpp, a tool for converting orthographic datasets to a consistent phonemic representation; and (ii) \ipachildes, a phonemic dataset of child-directed and child-produced speech across 31 languages. 
Prior tools for grapheme-to-phoneme conversion result in phonemic vocabularies that are inconsistent with established phonemic inventories, an issue which \gpp addresses by leveraging the inventories in the \phoible database \citep{phoible}. 
Using this tool, we augment CHILDES \citep{macwhinney1985child} with phonemic transcriptions to produce \ipachildes. 
This new resource fills several gaps in existing phonemic datasets, which often lack multilingual coverage, spontaneous speech, and a focus on child-directed language.
We demonstrate the utility of this dataset for phonological research by training phoneme language models on 11 languages and probing them for distinctive features, finding that the distributional properties of phonemes are sufficient to learn major class and place features cross-lingually.

\begin{tblr}{colspec = {Q[c,m]|X[l,m]}, stretch = 0}
    \cincludegraphics[width=1.4em, keepaspectratio]{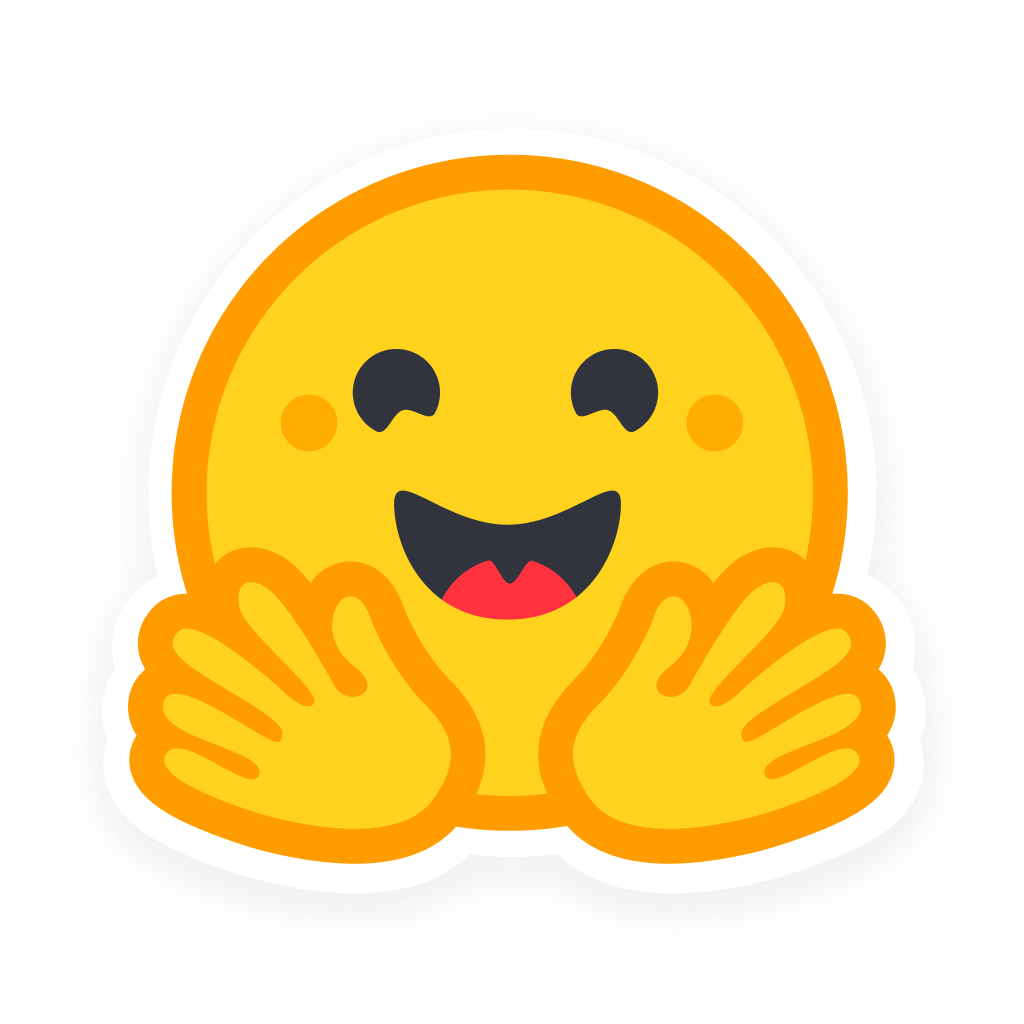} & {\footnotesize{\href{https://huggingface.co/collections/phonemetransformers/ipa-childes-67ee8533eb464db96ceb25b6}{phonemetransformers/ipa-childes}}\\ \tiny{(CC BY 4.0)} } \\
    \cincludegraphics[width=1.2em, keepaspectratio]{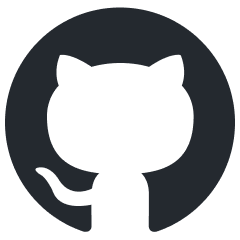} & {\footnotesize{\href{https://github.com/codebyzeb/g2p-plus}{codebyzeb/g2p-plus}}
    \tiny{(MIT)}}
\end{tblr}

\end{abstract}

\vspace{2mm}

\section{Introduction}

Phonological research can be enriched by large-scale data-oriented studies that investigate phoneme function across the globe's languages. However, while written text is plentiful and easily accessible across hundreds of languages, phonemic data is much more limited in availability. Phonemic datasets can be created by employing expert phoneticians to carefully transcribe speech, but this is a time-consuming process and completely infeasible for creating large datasets. Instead, the typical approach is to use grapheme-to-phoneme (G2P) conversion tools, which use statistical rules and pronunciation dictionaries to convert orthographic text to a phonemic representation. Open-source G2P tools have been used to create large and multilingual phonemic datasets with domains ranging from telephone conversations to legal proceedings. However, the fact that these tools are open-sourced and use a variety of statistical approaches and transcription schemes means that phonemic corpora vary considerably according to their phonemic vocabularies and level of phonetic detail, making it difficult to compare findings and incorporate other linguistic resources into analysis.

\begin{figure}[t]
    \centering
    \includegraphics[width=0.99\linewidth]{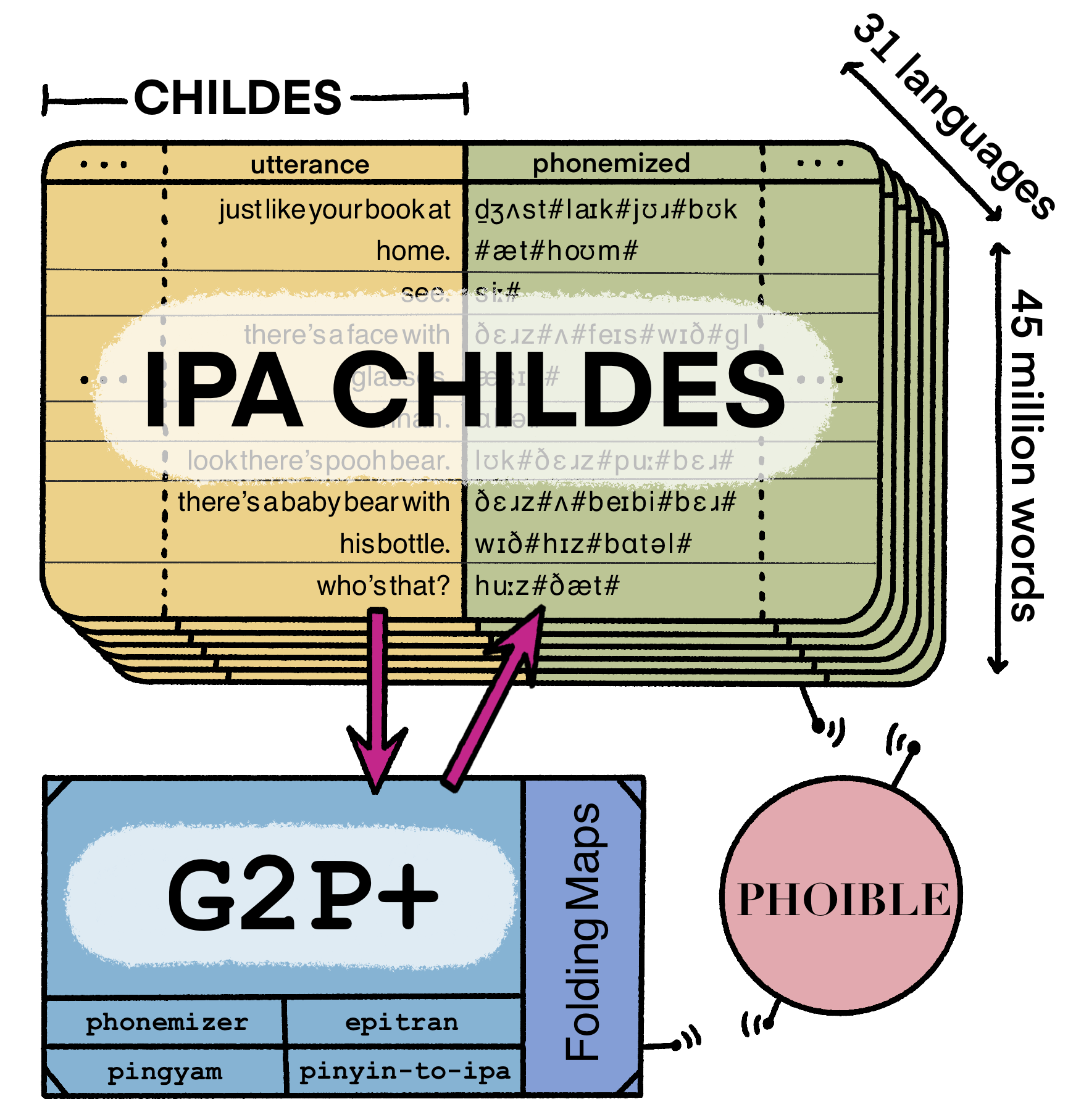}
    \caption{An overview of 
    \ipachildes and \gpp, which are introduced in this paper.}
    \label{fig:overview}
\end{figure}

There is also a lack of phonemic data for certain domains, preventing phonological research in these areas. In particular, we note that it is difficult to find phonemic data for child-centered speech\footnote{Child-centered speech is speech occurring within a child's environment and includes child-directed and child-produced utterances.} and, in general, spontaneous speech across several languages. The \emph{de-facto} repository for child-centered data is the Child Language Data Exchange System (CHILDES), which currently contains over 1.4TB of transcript data in over 40 languages \citep{macwhinney1985child, macwhinney_understanding_2019}. The impact of CHILDES across clinical and linguistic research has been profound \citep{ratner2024augmenting} but the largely orthographic nature of the data has prevented phonological experimentation.\footnote{CHILDES does contain phonetic transcriptions for some languages as part of the PhonBank project, but only for a select few corpora and only for child-produced utterances, impeding the phonological analysis of child-\emph{directed} speech.} 

We thus identify two major challenges impeding phonological research. First, the lack of consistent G2P conversion, which we address by developing \gpp, a tool for converting orthographic text to a phonemic representation. \gpp leverages existing G2P tools for conversion but carefully maps the output to established phonemic inventories in \phoible, a database of cross-linguistic phonological inventory data. Using \phoible inventories not only ensures consistency for each language regardless of the G2P backend used, but the database also contains phonological feature information, supporting fine-grained phonological analysis. Second, we address the lack of a multilingual phonemic dataset of child-centered speech by using \gpp to convert the majority of the CHILDES database to phonemes. The resulting dataset, \ipachildes, contains phonemic transcriptions of 31 languages in CHILDES, totaling 45 million words. We illustrate these resources in \cref{fig:overview}.

We exemplify how to use these resources by training cross-lingual phoneme language models. Phoneme LMs have a wide variety of applications in NLP, including lyric generation \citep{ding-2024-songcomposer}, text-to-speech \citep{li-2023-phoneme-level-bert}, and low-resource language modeling \citep{leong-whitenack-2022-phone}. Developmentally plausible training corpora also provide a means of studying emergent phonology, but past work has been limited by the availability of training and evaluation resources in languages besides English. Here, after establishing the scaling conditions of phoneme LMs, we train monolingual models on the 11 largest languages in \ipachildes. Using the fact that \gpp maintains a correspondence with \phoible during conversion, we use linear probes to predict an input phoneme's phonological features from its contextual embedding. We evaluate this approach against the phoneme's feature description in \phoible and find that the probes consistently correctly predict the `syllabic' and `consonantal' features, indicating the broad separation of vowels and consonants across languages and demonstrating the utility of phoneme LMs for studying emergent phonology.

These experiments demonstrate the utility of our tools for phonological analysis. We release \gpp, \ipachildes, and all trained models to support future work.

\section{Related Work}

\subsection{Phonemic Datasets}\label{sec:phonemicdatasets}

Phonemic data is required to investigate a range of linguistic phenomena. Recently, researchers have used data-driven approaches to study morphological theories of acquisition \citep{kirov2018recurrent}, explore the role of distributional information in phonology \citep{mayer-2020-phonology-distribution}, calculate cross-language phonological distance \citep{eden-2018-phonological-distance} and simulate early lexicon learning \citep{goriely2023word}. Despite the benefits of phonemic data, few such datasets exist.

Written text and audio datasets are far more plentiful than phonemic datasets. Written text, being widely distributed and easy to collect through practices such as web-scraping \citep{bansal-2022-datascaling}, has steered years of NLP research, ranging from the parsers trained on the Penn Treebank \citep{taylor2003penn} to the large language models trained on billion-word datasets like the Pile \citep{pile}. Despite the availability of written text, it is often inappropriate for speech technology and phonological research. Instead, since tape recorders became widely available, researchers have created datasets of human speech. These now include \textbf{elicited} speech corpora such as TIMIT, \citep{garofolo1993darpa}, FLEURS \citep{conneau2023fleurs}, the MSWC \citep{mazumder2021multilingual}, GlobalPhone \citep{schultz2002globalphone} and CommonVoice \citep{ardila-etal-2020-common}; \textbf{audio book} corpora such as LibriSpeech \citep{panayotov2015librispeech}, MLS \citep{pratap2020mls} and the CMU Wilderness Corpus \citep{8683536}; and \textbf{naturalistic} speech corpora such as Switchboard \citep{godfrey1992switchboard}, the Fisher corpus \citep{cieri2004fisher}, the British National Corpus \citep{bnc2007}, the Buckeye corpus \citep{pitt2007buckeye}, Babel \citep{harper2011babel} and VoxLingua107 \citep{9383459}. Of these datasets, only TIMIT, MLS and Switchboard include phonemic annotations, limiting their use in phonological analysis. Later work augmented these datasets with phonemic transcriptions. These include Audio BNC derived from the British National Corpus \citep{coleman2011mining}, LibriLight derived from LibriSpeech \citep{Kahn_2020}, VoxClamantis derived from the CMU Wilderness Corpus \citep{salesky-etal-2020-corpus}, VoxCommunis derived from CommonVoice \citep{ahn-chodroff-2022-voxcommunis} and IPAPACK derived from FLEURS and MSWC \citep{zhu-etal-2024-taste}.

These datasets and their phonemically-annotated successors all vary considerably according to the language coverage, number of words, domain and the presence of text-based transcriptions. We provide a summary of these properties in \cref{app:datasetstats}. Our dataset, \ipachildes, is the first phonemic dataset for \emph{child-centered} speech and the first \emph{multilingual} phonemic dataset for spontaneous speech.
\subsection{Grapheme to Phoneme Conversion}

Ideally, phonemic transcriptions of speech would originate from expert human annotators, but such annotation is incredibly slow. For instance, it was estimated that it would take 120 person-years to transcribe and align the 1200 hours of speech in the Audio BNC corpus \citep{coleman2011mining}. Of the phonemic datasets described above, only the smallest, TIMIT, was fully transcribed by human experts, at a rate of only 100 sentences per week \citep{zue1996transcription, lamel1989speech}. Switchboard also provides human-annotated phonemic transcriptions but only for 5,000 utterances \citep{greenberg1996insights}.

In practice, phonemic transcriptions are produced using G2P. In the simplest case, this involves the use of pronunciation dictionaries such as the Carnegie Mellon University (CMU) Pronouncing Dictionary\footnote{\url{http://www.speech.cs.cmu.edu/cgi-bin/cmudict}} or the English Pronouncing Dictionary \citep{jones2011cambridge}. These were used to create the phonemic transcriptions for the Buckeye Corpus, Audio BNC and Babel, but pronunciation dictionaries are limited by the items included in the dictionary and so may fail to convert part-words, interruptions or rare proper nouns, which frequently occur in spontaneous speech. More sophisticated G2P methods combine pronunciation dictionaries with statistical models. These systems have been developed for many languages using rules or finite-state transducers to generalize to unseen words \citep{Mortensen-et-al:2018, johnson2020g2p, Bernard2021}. Other G2P systems have applied neural networks to automatically learn these rules and generalize to new languages \citep{NOVAK_MINEMATSU_HIROSE_2016, Zhu2022}.

\begin{figure}[t]
    \centering
    \includegraphics[width=0.8\linewidth]{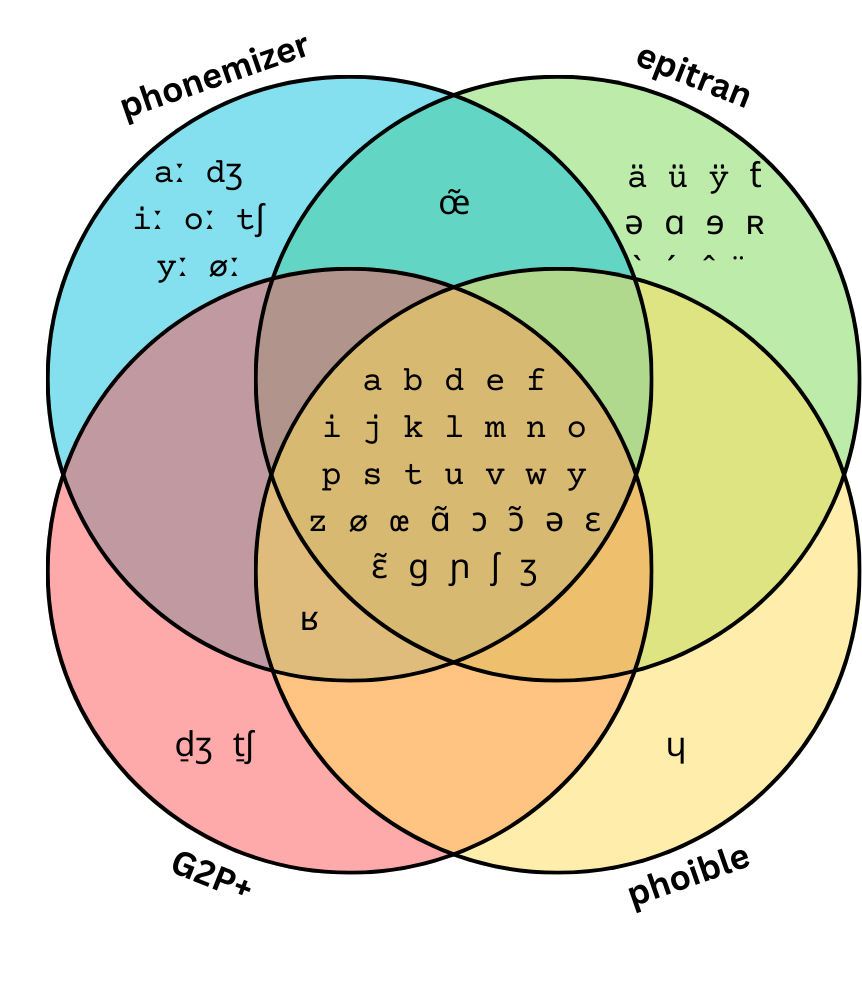}
    \caption{Venn diagram of the inventories produced by \texttt{phonemizer}, \texttt{epitran} and \gpp compared to \phoible inventory \href{https://phoible.org/inventories/view/2269}{2269} for French.}
    \label{fig:venn}
\end{figure}

As G2P systems operate only from text, they may fail to capture accents and the variation found in natural speech (see \cref{sec:limitations} for a discussion). Nevertheless, G2P systems provide a useful method for producing phonemic transcriptions at scale, and were used to produce the transcriptions for LibriSpeech, VoxClamantis and IPAPACK. The fact that transcription errors may occur is often acknowledged as a limitation, but rarely are the outputs of different G2P systems compared to each other or to established inventories. For instance, \texttt{epitran} and \texttt{phonemizer}, two popular tools described in \cref{sec:backends}, produce very different inventories for French, as demonstrated in \cref{fig:venn}. 

In this work, we leverage existing statistical G2P tools, validate their outputs using maps to \phoible inventories, and use our resulting tool to produce phonemic transcriptions for the utterances in the CHILDES database.

\subsection{Phoneme LMs and Child-Centered Data}\label{sec:babylm}

In this work, we illustrate one use of our dataset by training small monolingual LMs on 11 languages and examining the representations they learn for individual phonemes. 

Training models on such little data (here, only 500 thousand words) may be considered atypical in the modern NLP landscape, but questions of developmental plausibility have led to an increased interest in pretraining with limited data. For instance, the BabyLM workshop series challenges participants to train smaller models on data that is limited by both scale, 10--100 million words, and by domain, with the pre-training corpus including data from CHILDES, among other child-centered corpora \citep{warstadt-2023-babylm-findings, hu-etal-2024-findings}. Such limitations have led to the development of new architectures \citep{georges-gabriel-charpentier-samuel-2023-layers, charpentier2024gpt}, motivated cognitively-inspired pre-training strategies \citep{huebner-etal-2021-babyberta, martinez-etal-2023-climb} and allowed for gaining insights into human learning \citep{yedetore-etal-2023-poor}. The majority of this work has centered on English. Exceptions include \citet{capone2024babies, shen2024bambino}, who train Italian monolingual and bilingual models, respectively, \citet{yadavalli2023slabert} who use data from five language in CHILDES to explore second language acquisition theories (but only train an English LM) and \citet{salhan-etal-2024-less}, who use age-ordered data from four languages in CHILDES to explore fine-grained curricula inspired by language acquisition.

However, these BabyLMs are typically trained on orthographic text, limiting their ability to be studied at the phonological level, and generally use subword tokens, which do not generally correspond to cognitively plausible units \citep{beinborn-pinter-2023-analyzing} limiting their value for psycholinguistic research \citep{giulianelli-etal-2024-proper}. \citet{bunzeck2024graphemes} and \citet{goriely2024babble} both establish phoneme-based training of BabyLMs (where tokens consist of individual phonemes, with word boundaries removed) but only train on English text. Here, we use \ipachildes to demonstrate phoneme-based training for 11 languages and leverage the fact that \gpp maintains a correspondence to \phoible in order to probe our BabyLMs for knowledge of distinctive features. 

\section{\gpp}

We introduce \gpp as a tool for converting datasets from an orthographic representation to a phonemic representation. It operates either as a python library or as a command-line program; the user selects one of four backends and the language to use for conversion. Each backend supports a different set of languages as described in \cref{sec:backends}. The recommended backends for each of the languages in \ipachildes are given in \cref{app:breakdown} and example usage of the tool is given in \cref{app:usage}.

Each line of orthographic text is converted to phonemes, represented using the International Phonetic Alphabet (IPA). Regardless of the backend selected, the representation is consistent, with phonemes separated by whitespace (for convenient tokenization) and unique delimiters used to separate words and utterances (see \cref{sec:phonemestream} for details). 

The output representation is also consistent in terms of the set of phonemes types produced, using \emph{folding}, as described in \cref{sec:folding}. Without folding, each backend produces a different set of phonemes (as demonstrated in \cref{fig:venn}) which may not align with established phoneme inventories. Our folding maps not only ensure the output is consistent regardless of the backend chosen, but also makes it easy to leverage information in \phoible in analysis, as demonstrated in \cref{sec:featureprobing}.

\subsection{G2P Backends}\label{sec:backends}

In order to support a wide variety of languages, we implement wrappers around four backend G2P tools:

\paragraph{\texttt{phonemizer}:} Phonemizer \citep{Bernard2021} is a python library for G2P in various languages based on \texttt{eSpeak}\footnote{For Japanese text written in Romanji, as is the case in CHILDES, we use phonemizer with the the Segments backend \citep{robert_forkel_2019_3549784}.}, an open-source speech synthesizer which supports over one hundred languages and accents \citep{Dunn2019}.

\paragraph{\texttt{epitran}:} Epitran \citep{Mortensen-et-al:2018} supports the automatic grapheme-to-phoneme conversion of text across many languages, accents and scripts, with a particular focus on low-resource languages. For the majority of the 92 languages supported,\footnote{For English, Epitran uses the Flite Speech Sythesis System \citep{black2001flite} and for Simplified and Traditional Chinese it uses the CC-CEDict dictionary (\url{https://cc-cedict.org}).} it uses greedily-interpreted grapheme-to-phoneme maps augmented with context-sensitive pre-processor and post-processor rewrite rules.

\paragraph{\texttt{pinyin-to-ipa}:} Pinyin-to-ipa \citep{taubert_2024_pinyin-to-ipa_2024} is a python library for converting Mandarin written in pinyin to IPA using a few contextual grapheme-to-phoneme maps. The phoneme inventory is based on the phonology of Mandarin as described by \cite{lin2007sounds} and \cite{duanmu2007phonology} and tone markers are attached to the vowel of the syllable, rather than the end of the syllable. The tool only converts individual pinyin syllables, so our wrapper first splits the input into syllables before using the tool to convert each syllable to IPA.

\paragraph{\texttt{pingyam}:} Pingyam\footnote{\url{https://github.com/kfcd/pingyam}} is a table storing conversion information between the various romanization systems of Cantonese (including IPA) based on data from the Open Cantonese Dictionary.\footnote{\url{https://www.kaifangcidian.com/han/yue/}} Our wrapper converts from the Jyutping system to IPA by first splitting the input text into syllables before using the table to convert each syllable to IPA. For consistency with \texttt{pinyin-to-ipa}, we move tone markers to the vowel of each syllable. 

Although \texttt{pinyin-to-ipa} and \texttt{pingyam} only support one Chinese language each, we include them as backends because \texttt{epitran} and \texttt{phonemizer} have relatively poor G2P quality for these languages. This has prevented Chinese languages from being included in previous cross-lingual phonemic datasets \citep{ahn-chodroff-2022-voxcommunis} and has led to them being disregarded in cross-lingual analysis \citep{pimentel2020phonotactic}. We hope that by including these backends, we address this gap. We also combine tone markers with their preceding phoneme to create a unique token (e.g., \ttipa{\*a\tone{55}} is a single token, not two). We thus treat tone markers as phonological features rather than as individual phonemes, similar to how diphthongs are unique phonemes. However, this decision is still debatable and does lead to a comparatively larger phonemic vocabulary, so we provide an option to disable this merging (see \cref{app:usage}). 

\subsection{Phoneme inventory validation}\label{sec:folding}

In order to validate the set of phonemes produced by each choice of backend and language, we compare the output to the phoneme inventories for that language listed in \phoible, a database containing phoneme inventories extracted from source documents and tertiary databases for 2186 distinct languages \citep{phoible}.

\phoible also contains typological data and phonological feature information for each phoneme, a useful resource for phonological analysis. As there are often multiple inventories in \phoible for each language, we choose the inventory that best matches the output phoneme of all backends that supports that language, according to the number of phoneme types, the number of consonants, the number of vowels and the number of diphthongs.

Once the best inventory has been found, we use a process called \emph{folding} to align the output phoneme set with the inventory and correct errors in the output. This is achieved a manually-crafted look-up table (a \emph{folding map}) which is applied to the output of the G2P wrapper. These maps are primarily used to solve surface-level errors, instances where the G2P tool outputs a specific Unicode string for a specific phoneme but the inventory lists a different string. For example, the \texttt{phonemizer} backend with the \texttt{ja} language code (Japanese) outputs the tied characters \ttipa{\textroundcap{ts}} as one of the phonemes, but the Japanese inventory lists \ttipa{ts} instead. These errors can be solved with a simple one-to-one mapping. These mappings will not affect the information-theoretic properties of the output but do allow the output symbols to be matched with entries in \phoible.

Besides these surface-level errors, other transcription errors can also be solved with folding maps. For example, the \texttt{epitran} backend for Serbian always outputs \ttipa{d Z} as two phonemes instead of the single phoneme \ttipa{dZ}, which can also be solved with a single mapping. The construction of the folding maps and these additional error types are discussed further in \cref{sec:folding-details}. 

\subsection{Qualitative Analysis}\label{sec:qualitative}

In \cref{fig:venn}, we compare the matching \phoible inventory for French to the output of \gpp (using \texttt{phonemizer} as a backend) and the outputs produced by \texttt{phonemizer} and \texttt{epitran} when applied to the French section of CHILDES. The outputs of \texttt{phonemizer} and \texttt{epitran} both differ considerably from the inventory and from each other whereas the \gpp only fails to produce a single phoneme, \ttipa{\textturnh}, and produces two additional phonemes \ttipa{dZ} and \ttipa{tS}, which we allow as they come from loanwords such as ``pizza'' and ``sandwich''. 

\section{\ipachildes}

\ipachildes contains 45 million words of monolingual child-centered speech for 31 languages. The data is sorted by child age in order to support curriculum learning experiments, such as in the work of \citet{huebner-etal-2021-babyberta}, and we also provide an `is\_child' feature to allow for filtering child or adult utterances.

In order to create the dataset, we first download all monolingual and non-SLI corpora in CHILDES.
CHILDES has 48 languages but only 31 are supported by a backend in \gpp (either because the language is not supported, or because they have been transcribed using an irregular script). For languages supported by multiple backends, we produce a sample transcription using each backend and carefully examine the output. The `best-fitting' backend (the one that produces a phonemic vocabulary closest to one of the inventories in \phoible) is selected and is the backend for which we produce a folding map, as described in \cref{sec:folding}. Having selected the best backend, we use \gpp to convert all orthographic utterances for each language to a phonemic representation, producing a CSV containing the original representation, the phonemic representation as well as additional data stored in CHILDES (such as target child age, morpheme count, part of speech information, and the IDs of each utterance, transcript, corpus and collection). 

An illustration of the dataset is given in \cref{fig:overview} and a description of each language section is given in \cref{app:breakdown}, detailing the matching \phoible inventory and CHILDES section for each language. Note that English is divided into British English (EnglishUK) and North American English (EnglishNA) to mirror the split present in CHILDES and Portuguese is also split into European and Brazilian varieties, following previous work \citep{caines2019cross, goriely2023word}. For these splits, we use different \texttt{phonemizer} accents. Data is not uniformly distributed across languages. EnglishNA is the most represented, with close to 10 million words, and Farsi is the least represented, with only 43 thousand words. We discuss limitations of the dataset in \cref{sec:limitations}.

\section{Cross-Lingual Phoneme LMs}\label{sec:phonemelms}

\begin{figure*}[t]
\centering
\begin{subfigure}{0.45\linewidth}
    \includegraphics[width=\linewidth]{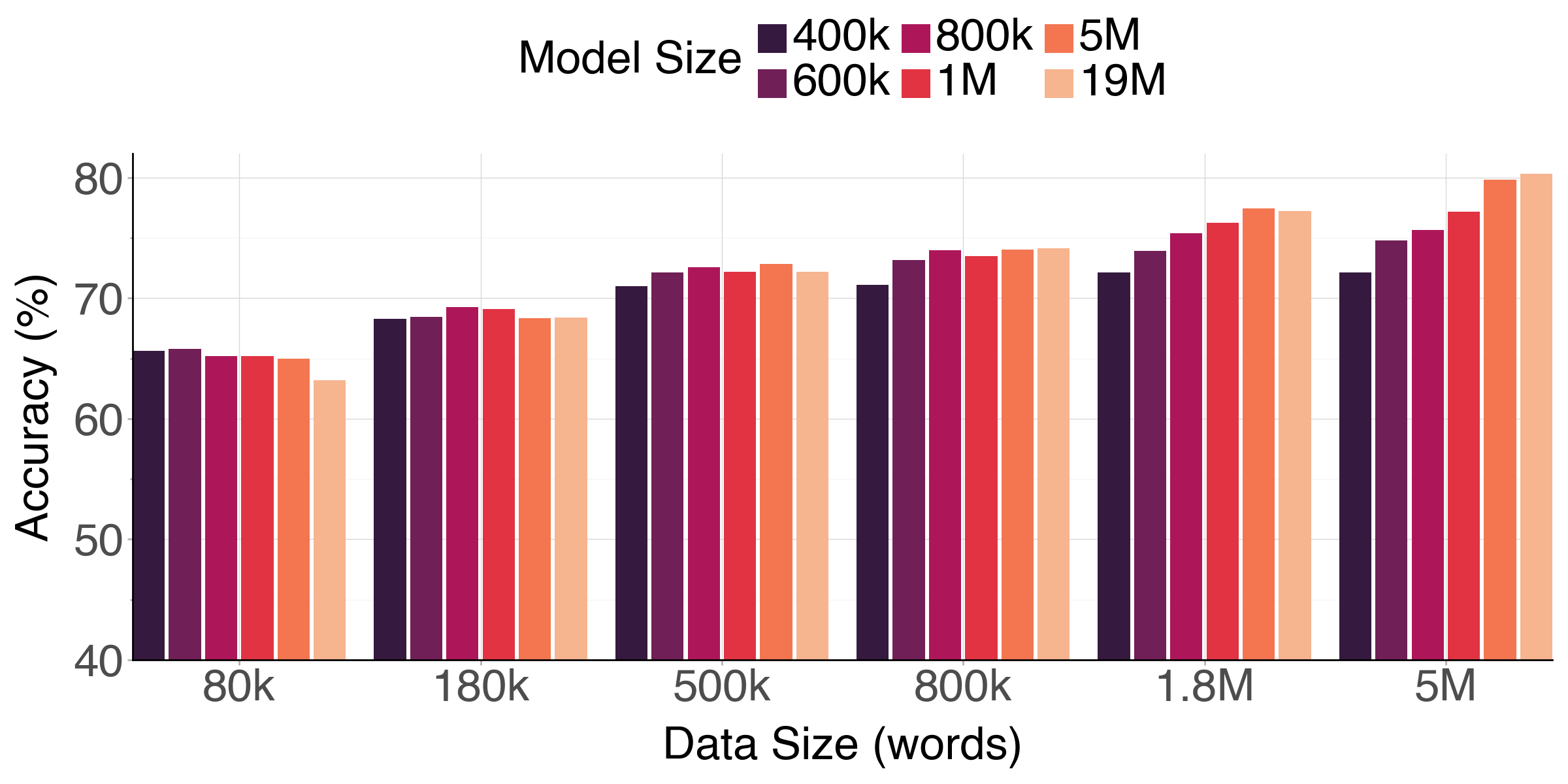}
\end{subfigure}
\hfill
\begin{subfigure}{0.45\linewidth}
    \includegraphics[width=\linewidth]{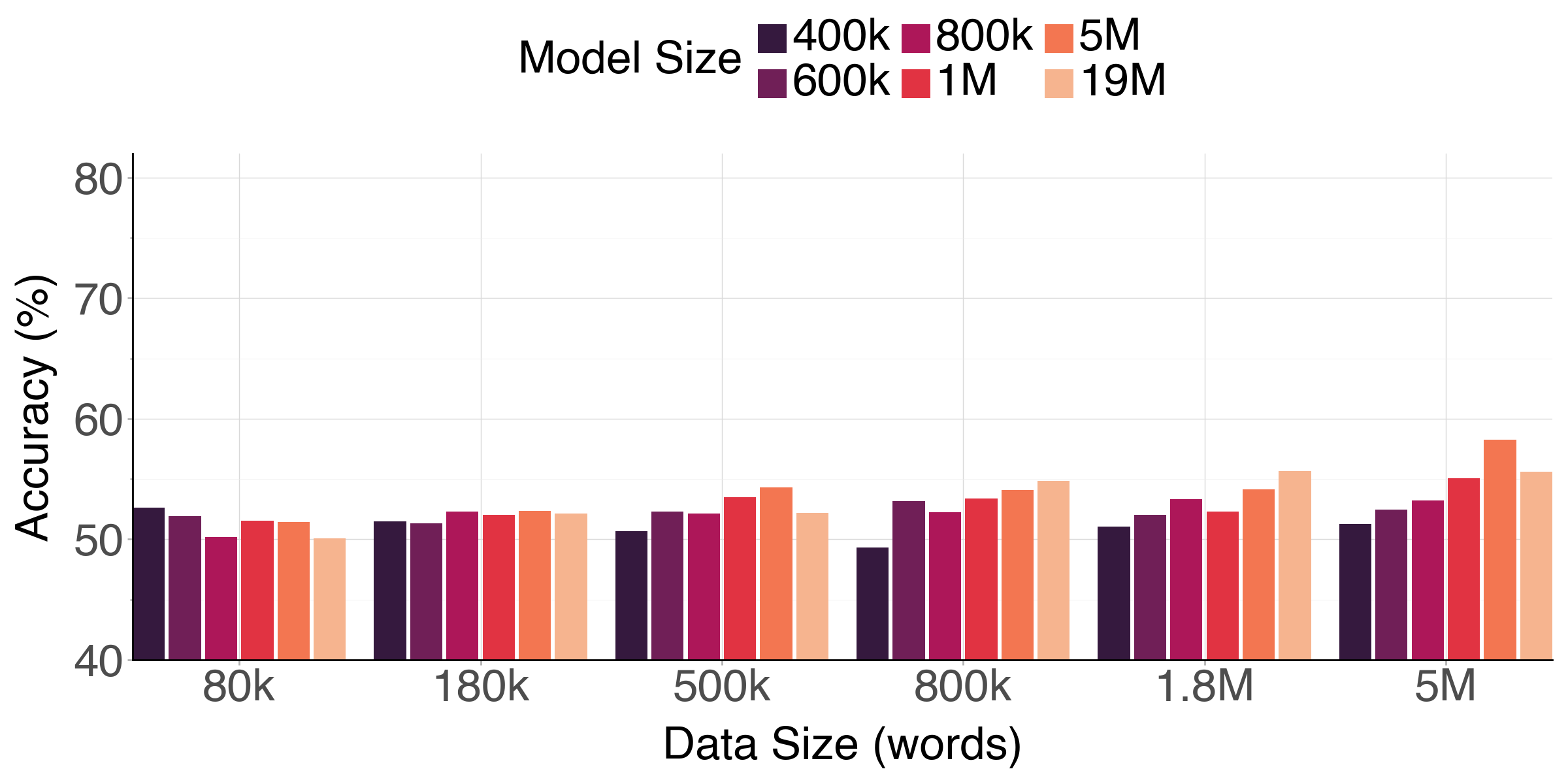}
\end{subfigure}
\hfill
\caption{BabySLM lexical score (left) and syntactic score (right) achieved by a phoneme-based GPT-2 model trained on the EnglishNA portion of \ipachildes across model sizes and subsample sizes.}
\label{fig:babyslm}
\end{figure*}

Phoneme LMs trained on developmentally plausible corpora allow for the testing of phonological representations but recent work has only explored English models trained on 10 -- 100 million words (see \cref{sec:babylm}). Here, we establish the size requirements for models trained on data available in \ipachildes and then demonstrate how models trained on the 11 largest languages in our dataset can be used to explore emergent phonology.

Each of our models are auto-regressive, trained to predict phonemes in a sequence. This is similar to how standard auto-regressive models are trained, except that each token represents a single phoneme, rather than a word or subword. We refer to the suite of models as ``cross-lingual'' as each individual model is monolingual, only trained on data from a single language. This is in contrast to ``multilingual'' models that are trained on multiple languages at once.

\subsection{Size Requirements of Phoneme LMs}\label{sec:sizerequirements}

We use the BabySLM benchmark \citep{lavechin} to evaluate syntactic and phonological knowledge. The \emph{syntactic} score is calculated using a preference task over pairs of grammatical and ungrammatical sentences across six syntactic phenomena commonly seen in naturalistic speech. For example, models should assign \texttt{\textipa{D~@~g~U~d~k~I~t~i}} (``the good kitty'') a higher likelihood than \texttt{\textipa{D~@~k~I~t~i~g~U~d}} (``the kitty good''). The \emph{lexical} score is similarly calculated using minimal pairs of words and pseudo-words, such as \texttt{\textipa{\*r~u:~l~@~\*r~z}} (``rulers'') compared to the pseudo-word \texttt{\textipa{\*m~u:~k~@~\*r~z}} (``mukers''). \citet{lavechin} demonstrated that an LSTM model trained on 1.2 million words from Providence (one of the corpora in CHILDES) achieved a lexical score of 75.2 and a syntactic score of 55.1\footnote{Chance performance for both BabySLM scores is 50 and 100 indicates perfect performance}. \citet{goriely2024babble} later achieved lexical and syntactic scores of 87.8 and 83.9 when training a larger transformer-based model on the 100-million-word BabyLM challenge dataset \citep{conll-2024-babylm}.

Here, we use \ipachildes and BabySLM to establish the scaling laws of phoneme LMs in terms of data size and model size. We subsample the EnglishNA portion of the dataset, remove word boundaries and child-produced utterances and train a suite of GPT-2 models ranging from 400 thousand to 19 million non-embedding parameters. To prevent overfitting, we train three models for each combination of model size and data size using dropouts of 0.1, 0.3 and 0.5, selecting the model with the lowest perplexity for each. Model parameters, training configurations and scripts are provided in \cref{sec:models}.

The scaling graphs for the lexical and syntactic scores are given in \cref{fig:babyslm}. For every model size, performance increases with more training data but for a particular data size the largest model is not always the best. For instance, the second smallest model is the best choice for the lexical task if only 300 thousand tokens of data are available, likely due to larger models overfitting with a sample this small (even with high dropout). It is also clear that although small models with very little data seem to acquire phonological knowledge (as measured by the lexical score), much more data is required to achieve syntactic scores past 60, in line with the results of \citet{lavechin} and \citet{goriely2024babble}. The best model parameters for each score and data size are given in \cref{sec:best-model-parameters}.

\subsection{Probing for Phonological Features}\label{sec:featureprobing}

As the phonemic utterances in \ipachildes maintain a correspondence with \phoible, we can use the \textbf{distinctive feature} information in \phoible to probe cross-lingual phoneme LMs for phonological knowledge. 

We select the 11 largest languages in the dataset and train a GPT-2 model on each, subsampling 500 thousand words\footnote{As the number of phonemes per word varies across these languages, we actually subsample 1.8 million tokens (phonemes) for each language, which is roughly 500 thousand words.} and using the best-fitting model for this data size according to the previous experiment (the 5-million-parameter model with a dropout of 0.3). The training configuration remains the same (see \cref{sec:models}). These models allow us to compute contextual embeddings $c(x)$ for phonemes.

\begin{figure}[t]
    \centering
    \includegraphics[width=0.99\linewidth]{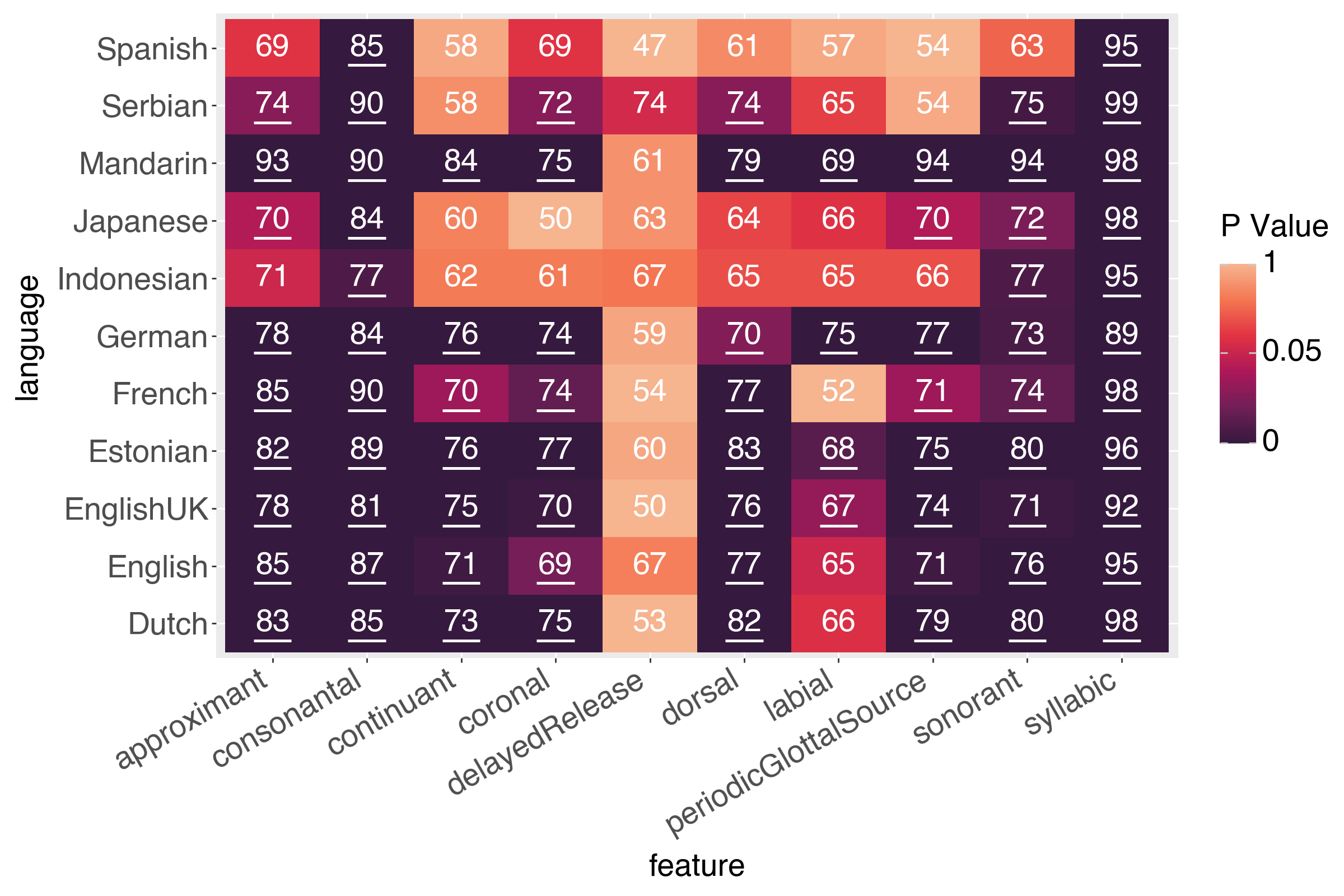}
    \caption{Accuracy of the phonological distinctive feature probe across 11 languages in \ipachildes and 9 distinctive features from \phoible.}
    \label{fig:features}
\end{figure}

\begin{figure}[t]
    \centering
    \includegraphics[width=0.99\linewidth]{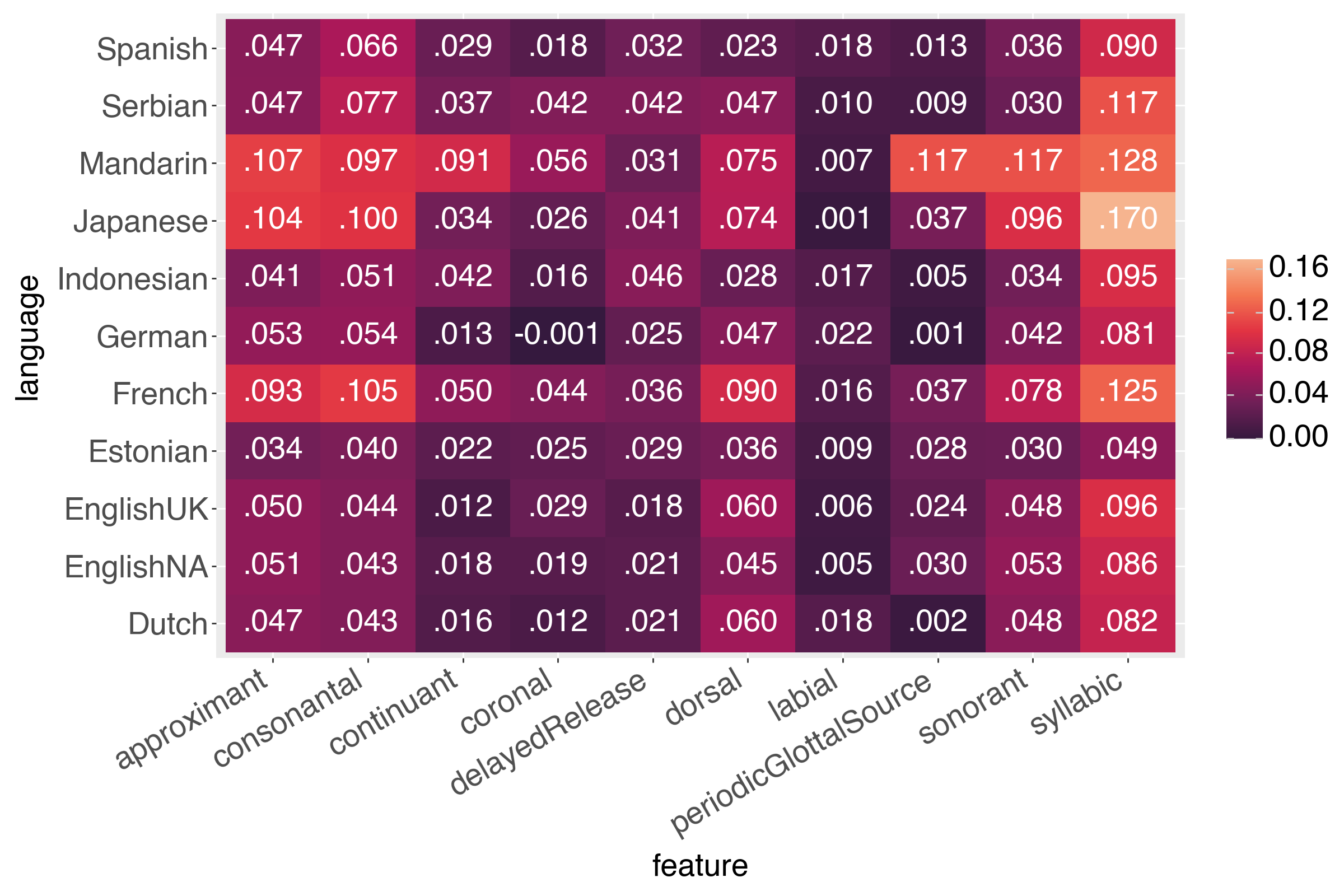}
    \caption{Average silhouette scores when using each distinctive feature to cluster contextual embeddings of the phonemes in each language.}
    \label{fig:silhouette}
\end{figure}

\begin{figure*}[t]
    \centering
    \begin{subfigure}[][][]{0.49\textwidth}
        \includegraphics[width=0.99\linewidth]{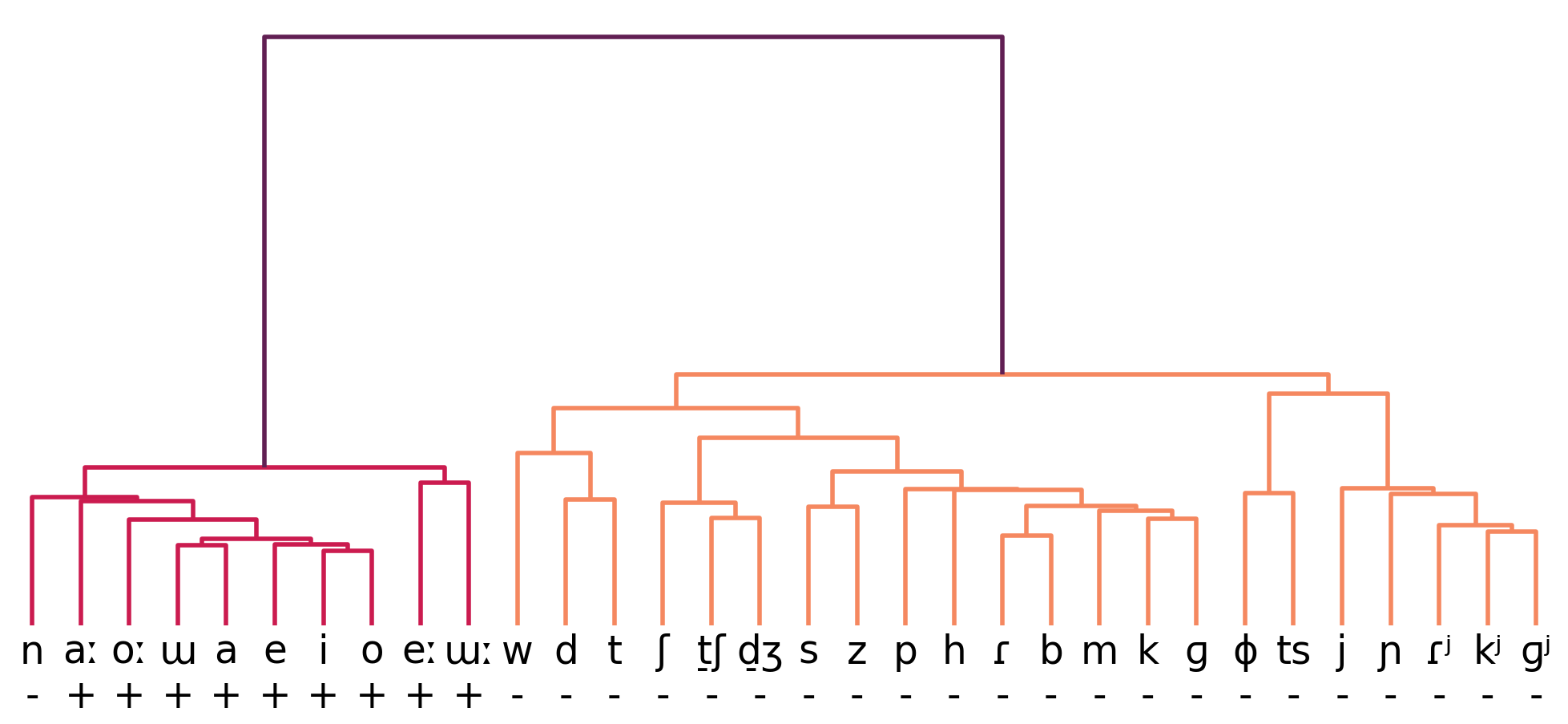}
        \caption{Japanese}
    \end{subfigure}
    \begin{subfigure}[][][]{0.49\textwidth}
        \includegraphics[width=0.99\linewidth]{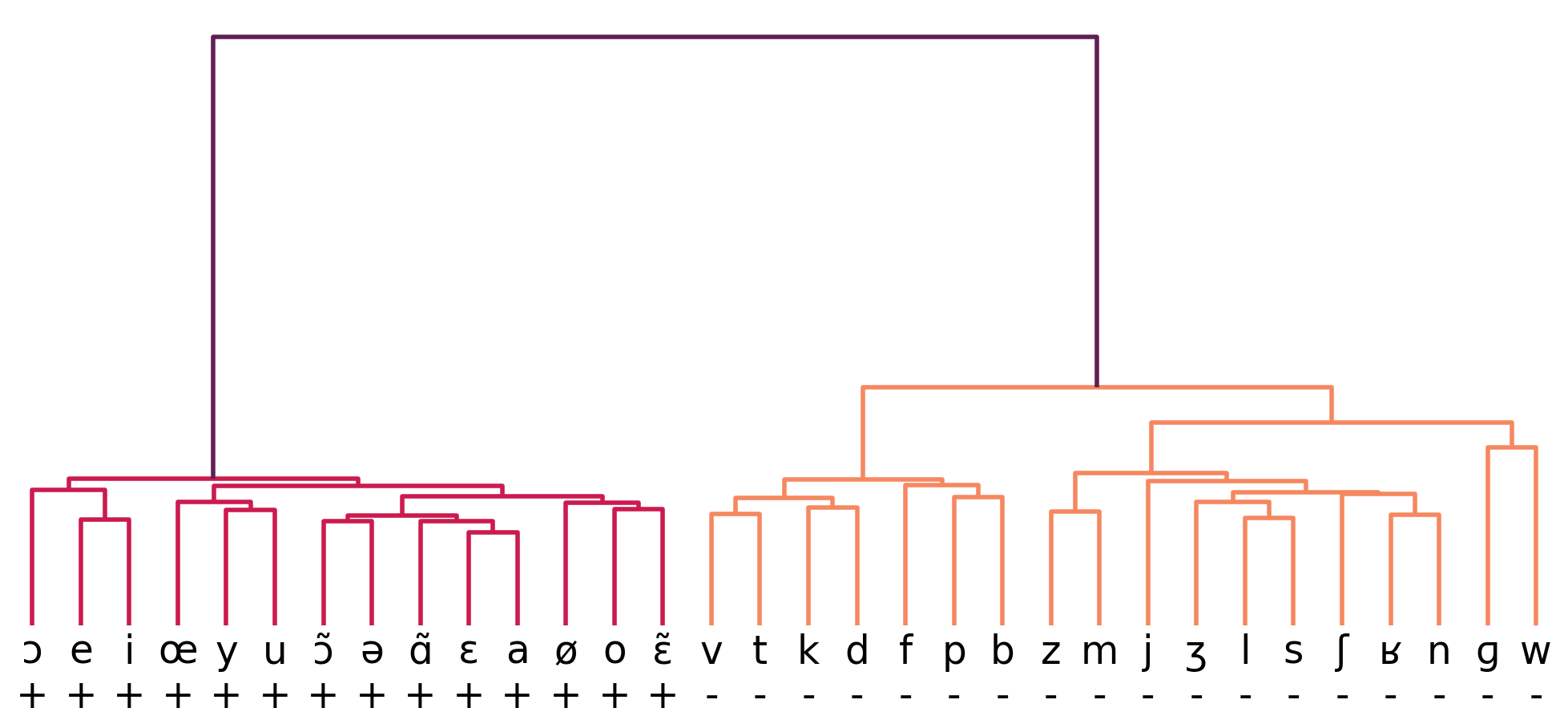}
        \caption{French}
    \end{subfigure}
    \caption{Similarity of the contextual embeddings for each phoneme learned by the Japanese and French phoneme LMs. Similarities are computed using Euclidean distance considering the average of 50 contextual embeddings for each phoneme and linkages are created using the incremental algorithm. The `syllabic' distinctive feature is marked below each phoneme.} 
    \label{fig:dendrogram}
\end{figure*}

We then look up the distinctive features of each phoneme in each language using the matching inventories in \phoible (see \cref{tab:dataset-phonemized-childes-sections}). We find the set of features for which, in all 11 languages, there are at least 4 phonemes that exhibit the feature and 4 that do not. 
For each feature $f$, we train a linear probe $p_f$ to predict that feature from the contextual embeddings $c(x)$ of phonemes. Each probe is trained with an equal number of positive and negative examples and is evaluated using leave-one-group-out cross-validation (i.e for each phoneme $x$ in the phoneme inventory $V$, the probe is trained on the contextual embeddings of all other phonemes $\{c(y) | y\in V \setminus \{x\}\}$, then evaluated by predicting the feature from contextual embeddings of the left-out phoneme $p_f(c(x))$, and the final score is a macro-average across all phonemes $x\in V$).

The results of each probe are provided in \cref{fig:features}. The majority of the probes achieve accuracies significantly\footnote{Statistical significance was assessed using a binomial test, where the null hypothesis assumes a probability of success \( p_0 = 0.5 \) and the number of trials \( n \) is equal to the number of phonemes tested by the probe. A result was considered significant if the computed \( p \)-value was less than 0.05.} higher than chance (50\%), indicating that the models learn representations that encode distinctive features. While the scores for each feature are broadly consistent across languages, some notable differences emerge. For example, nearly all feature probes achieve statistically significant results in Mandarin, whereas only two do so in Spanish. This disparity can be partly attributed to the number of unique phonemes in each language. Because we treat each combination of vowel and tone as a distinct phoneme, Mandarin has 99 phoneme types, compared to just 24 in Spanish. The smaller phoneme inventory in Spanish greatly reduces $n$ for each probe, making it more challenging to obtain statistically significant results.

In all 11 languages, the highest result is achieved by the probe for the `syllabic' feature which generally\footnote{In some languages there are also syllabic consonants, which like vowels can act as the nucleus of a syllable.} separates vowels from consonants. As these models only learn to predict phonemes and have no concept of how each phoneme is pronounced, the fact that this separation is learned clearly indicates that vowels and consonants provide a strong distributional signal across languages. The \texttt{consonantal} feature similarly separates vowels from consonants\footnote{This feature indicates an audible constriction of the vocal tract, separating obstruents, nasals, liquids, and trills from vowels, glides and laryngeal segments \citep{gussenhoven2017understanding}.} and is learned by a probe in every language. However, not every feature can be learned by these probes. For instance, the \texttt{delayedRelease} feature, which distinguishes stops from affricates, is not learned by any probe. Our models do not encode the rate of phoneme delivery, so it is unsurprising that a feature that relates to the temporal properties of phonemes is difficult to probe.

\subsubsection*{Distributional Phoneme Clusters}

To better understand why the probes capture certain phonological features, we examine whether contextual embeddings cluster according to these features. For each language, we sample 50 contextual embeddings per phoneme and label them with their associated phonological features. For each labeling, we then compute the \textbf{silhouette score} for each embedding --- a metric ranging from –1 to 1, where higher values indicate that an embedding is more similar to others in its assigned cluster than to those in neighboring clusters \citep{rousseeuw1987}. Averaging these scores across all embeddings allows us to compare how well different features cluster the phoneme representations, as shown in \cref{fig:silhouette}.

The scores are all relatively close to zero, likely due to the curse of dimensionality --- our embeddings have 256 dimensions, far exceeding the number of distinct phonemes in each language. Despite this, the results are consistent with the probe findings: the syllabic feature yields the highest clustering quality.

We further visualize this clustering using dendrograms, created by averaging the contextual embeddings for each phoneme and applying an incremental clustering algorithm. \Cref{fig:dendrogram} shows examples for Japanese and French, with the syllabic feature marked for each phoneme. In both cases, vowels are almost entirely separated from consonants, with one notable exception: \ttipa{n} in Japanese. We also observe some alignment with traditional phoneme groupings (e.g., \ttipa{b} and \ttipa{p}), though overall the dendrograms diverge from standard phonological classifications. This suggests that the distributional behavior of phonemes in context may not neatly align with their articulatory or categorical properties.

\section{Discussion}

\ipachildes addresses several limitations of past datasets, as the first large multilingual corpus of child-centered phonemic speech. In this study we demonstrate how this data can be used to train phoneme LMs, but this dataset could also support information-theoretic studies of language processing and acquisition, which have previously based their calculations on word types \citep{piantadosi2011word, dautriche2017words, pimentel2020phonotactic} or orthographic text \citep{mahowald2013info, dautriche2017wordform, futrell2020lossy}, often citing a lack of phonemic data as a limiting factor.
The child-centered domain of our dataset could also be beneficial for studying the `Goldilocks' hypothesis \citep{kidd2014goldilocks} and the properties of `Parentese' \citep{ramirez2017look}. We provide an example of an experiment investigating the later in \cref{app:parentese}, where we compute the average information of utterances directed to children aged 0--6 across 10 languages and find a general trend of increasing informative content.

Our \gpp tool also provides new avenues for linguistic analysis by ensuring that phonemes produced for each language are consistent with established inventories in \phoible. This not only addresses transcription errors, but also allows for the use of distinctive feature information provided by \phoible in analysis. We demonstrate this by training linear probes to extract distinctive features from the contextual embeddings of phonemes learned by our monolingual models. We find that certain features (e.g. \texttt{consonantal}) emerge solely from the distributional properties across all 11 languages, while others (e.g. \texttt{delayedRelease}) do not. 

Our resources could also support the training of self-supervised speech models \citep[e.g.][]{hsu2021hubert}. These models are trained directly on audio and lag behind phoneme or text-based models, often requiring several orders of magnitude more data to learn semantic representations \citep{cuervo2024scaling}, but recent work has found that fine-tuning on phoneme classification can reduce this gap \citep{feng-2023-language-universal-phonetic, poli2024improving}. Our work is closely related to recent efforts in low-resource cross-lingual language modeling --- for example, the Goldfish suite of monolingual models spanning 350 languages, some trained on as little as 5MB of orthographic text \citep{chang2024goldfish}.
IPA is also a more universal representation than orthographic text, which varies considerably across languages, with multilingual IPA models proving to be effective for force-alignment \citep{zhu-etal-2024-taste} and zero-shot cross-lingual NER \citep{sohn2024zero}. In this study we only train monolingual models, but future work could extend this to the multilingual setting.

\section{Conclusion}

This work introduces \gpp and \ipachildes, two new resources for phonological research. \gpp improves open-source G2P tools by ensuring phonemic vocabularies align with the established inventories in the \phoible database. Using this tool, we create \ipachildes by converting the orthographic transcriptions in CHILDES into phonemic representations, resulting in a large corpus of child-centered spontaneous speech across 31 languages.

We demonstrate the utility of these resources for phonological analysis using phoneme LMs by extending prior work to the cross-lingual setting. Our results establish the corpus size requirements for phoneme LMs trained on developmentally plausible corpora and we show that models trained on 11 languages effectively implicitely encode distinctive features. These findings support the role of phoneme LMs in studying emergent phonology. We anticipate that \gpp and \ipachildes will enable a wide range of future studies in linguistics and NLP, particularly in phonological acquisition, cross-linguistic analysis, and speech processing.

\section*{Acknowledgements}

We thank Lisa Beinborn for her advice on an an early draft of this article. We are also grateful to Pietro Lesci and Suchir Salhan for their feedback. 

Our experiments were performed using resources provided by the Cambridge Service for Data Driven Discovery (CSD3) operated by the University of Cambridge Research Computing Service, provided by Dell EMC and Intel using Tier-2 funding from the Engineering and Physical Sciences Research Council (capital grant EP/T022159/1), and DiRAC funding from the Science and Technology Facilities Council. Z\'ebulon Goriely is supported by an EPSRC DTP Studentship. 

\bibliographystyle{acl_natbib}
\bibliography{custom}

\newpage
\clearpage
\appendix

\section{Limitations}\label{sec:limitations}

We consider the following limitations of our work. 

\paragraph{Phonemes as a representation of speech:} While phonemic data more closely represents how words are pronounced compared to orthographic text (the degree of this difference varies between languages), phonemes are still abstract symbolic units which do not contain many of the detailed and continuous features of speech, such as prosody. They also abstract away from phones, which are detailed realizations of phonemes, representing the physical sound produced rather than a language-specific meaningful unit. When comparing modalities that may be close to the sensory signal available to infants for developmentally plausible language modeling, some researchers consider phonemic data to be as implausible as orthographic data \citep{lavechin} and instead create language models that can be trained directly on audio \citep{kamper2017segmental, nguyen2020zero, hsu2021hubert, dunbar2021zero}. Nevertheless, phonemes still provide a useful unit of analysis and are necessary for certain linguistic theories and information-theoretic calculations. While phones could offer another useful representation, they are even harder to source than phonemes.

\paragraph{G2P conversion inaccuracies:} Despite improving G2P conversion by mapping to inventories in the \phoible database, there are still limitations with \gpp. Firstly, our method integrates existing G2P tools, which may vary in quality between languages. When converting each language in CHILDES, we selected the most appropriate backend for each language, in particular adding two backends to support G2P for Mandarin and Cantonese, but the quality may still vary. Many of the G2P tools for certain languages convert words individually, so we do not capture vowel reduction, allophonic variation or other differences found in natural speech. We also use a single accent for each language, losing inter-speaker variability. The \texttt{phonemizer} backend supports multiple accents for certain languages (here we use a different accent for EnglishNA and EnglishUK) and future work could try to maintain accent differences during grapheme-to-phoneme conversion, but this would require speaker information or audio, as was done during the creation of Audio BNC \citep{coleman2012audio}. Finally, we note that G2P methods may not produce correct transcriptions for child-produced utterances, which are often corrected by the transcriber, especially for young infants. Initially we intended to distribute \ipachildes without child-produced utterances (and in this study only train models with the child-directed utterances) but as they might be useful in future research, we instead note this limitation.

\paragraph{Phoible inventories:} Although the \phoible database collects established phonemic inventories and provides distinctive feature vectors, there are still often multiple phoneme inventories for a single language. This the exact phonemic inventory for a particular language is still a matter of debate among expert phonologists. When creating folding maps we choose the `best-fitting' inventory to map to, as detailed in \cref{tab:dataset-phonemized-childes-sections}, but we acknowledge that these inventories may not be exact.

\paragraph{Phoneme LMs:} We train phoneme LMs on 11 languages from \ipachildes but the specific architecture we use is based on our scaling experiment for the EnglishNA model. Although we do not directly compare these LMs, we note the possibility that other parameters may have better suited the non-English languages. We were only able to conduct the scaling experiments for English due to the lack of phonological benchmarks for other languages but we hope that the release of \ipachildes facilitates further work in multilingual phonological evaluation of phoneme LMs.

\paragraph{Languages:} Although our dataset is multilingual, there are still limitations in terms of language coverage. The languages are predominantly European and Asian, with no languages indigenous to the Americas, Australia or Africa. English is also still the dominant language of the dataset and the Farsi section is very small, only containing 43 thousand words. In creating this dataset, we were limited by the languages available in CHILDES. The languages in CHILDES we were not able to convert were Greek, Arabic, Hebrew, Thai, Georgian, Tamil, Taiwanese, Jamaican, Sesotho, Berber, Cree and Slovenian and Russian due to missing G2P backends or unsupported orthographies.

\section{Breakdown of \ipachildes}\label{app:breakdown}

\ipachildes contains transcriptions of child-centered speech for 31 languages. Details of each language section are provided in \cref{tab:dataset-phonemized-childes-sections}.  

\begin{table*}[t]
    \centering
    \footnotesize
    \begin{tabular}{lllcccc}
        \toprule
        \textbf{Language} & \textbf{CHILDES Collection} & \textbf{Backend} & \textbf{Inventory ID} & \textbf{Words} & \textbf{Phonemes} & \textbf{\% Child} \\ 
        \midrule
        EnglishNA & \href{https://childes.talkbank.org/access/Eng-NA}{EnglishNA} (49) & \texttt{phonemizer} & \href{https://phoible.org/inventories/view/2175}{2175} & 9,993,744 & 30,986,218 & 36 \\
        EnglishUK & \href{https://childes.talkbank.org/access/Eng-UK}{EnglishUK} (16) & \texttt{phonemizer} & \href{https://phoible.org/inventories/view/2252}{2252} & 7,147,541 & 21,589,842 & 39 \\
        German & \href{https://childes.talkbank.org/access/German}{German} (10) & \texttt{epitran} & \href{https://phoible.org/inventories/view/2398}{2398} & 5,825,166 & 21,442,576 & 44 \\
        Japanese & \href{https://childes.talkbank.org/access/Japanese}{Japanese} (11) & \texttt{phonemizer} & \href{https://phoible.org/inventories/view/2196}{2196} & 2,970,674 & 11,985,729 & 44 \\
        Indonesian & \href{https://childes.talkbank.org/access/EastAsian}{EastAsian/Indonesian} (1) & \texttt{epitran} & \href{https://phoible.org/inventories/view/1690}{1690} & 2,347,642 & 9,370,983 & 34 \\
        French & \href{https://childes.talkbank.org/access/French}{French} (15) & \texttt{phonemizer} & \href{https://phoible.org/inventories/view/2269}{2269} & 2,973,318 & 8,203,649 & 40 \\
        Spanish & \href{https://childes.talkbank.org/access/Spanish}{Spanish} (18) & \texttt{epitran} & \href{https://phoible.org/inventories/view/164}{164} & 2,183,992 & 7,742,550 & 46 \\
        Mandarin & \href{https://childes.talkbank.org/access/Chinese}{Chinese/Mandarin} (16) & \texttt{pinyin\_to\_ipa} & \href{https://phoible.org/inventories/view/2457}{2457} & 2,264,518 & 6,605,913 & 39 \\
        Dutch & \href{https://childes.talkbank.org/access/DutchAfrikaans}{DutchAfricaans/Dutch} (5) & \texttt{phonemizer} & \href{https://phoible.org/inventories/view/2405}{2405} & 1,475,174 & 4,786,803 & 35 \\
        Polish & \href{https://childes.talkbank.org/access/Slavic}{Slavic/Polish} (2) & \texttt{phonemizer} & \href{https://phoible.org/inventories/view/1046}{1046} & 1,042,841 & 4,361,797 & 63 \\
        Serbian & \href{https://childes.talkbank.org/access/Slavic}{Slavic/Serbian} (1) & \texttt{epitran} & \href{https://phoible.org/inventories/view/2499}{2499} & 1,052,337 & 3,841,600 & 29 \\
        Estonian & \href{https://childes.talkbank.org/access/Other}{Other/Estonian} (9) & \texttt{phonemizer} & \href{https://phoible.org/inventories/view/2181}{2181} & 843,189 & 3,429,228 & 45 \\
        Welsh & \href{https://childes.talkbank.org/access/Celtic}{Celtic/Welsh} (2) & \texttt{phonemizer} & \href{https://phoible.org/inventories/view/2406}{2406} & 666,350 & 1,939,286 & 69 \\
        Cantonese & \href{https://childes.talkbank.org/access/Chinese}{Chinese/Cantonese} (2) & \texttt{pingyam} & \href{https://phoible.org/inventories/view/2309}{2309} & 777,997 & 1,864,771 & 34 \\
        Swedish & \href{https://childes.talkbank.org/access/Scandinavian}{Scandinavian/Swedish} (3) & \texttt{phonemizer} & \href{https://phoible.org/inventories/view/1150}{1150} & 581,451 & 1,782,692 & 45 \\
        PortuguesePt & \href{https://childes.talkbank.org/access/Romance}{Romance/Portuguese} (4) & \texttt{phonemizer} & \href{https://phoible.org/inventories/view/2206}{2206} & 499,522 & 1,538,408 & 39 \\
        Korean & \href{https://childes.talkbank.org/access/EastAsian}{EastAsian/Korean} (3) & \texttt{phonemizer} & \href{https://phoible.org/inventories/view/423}{423} & 263,030 & 1,345,276 & 37 \\
        Italian & \href{https://childes.talkbank.org/access/Romance}{Romance/Italian} (5) & \texttt{phonemizer} & \href{https://phoible.org/inventories/view/1145}{1145} & 352,861 & 1,309,489 & 39 \\
        Croatian & \href{https://childes.talkbank.org/access/Slavic}{Slavic/Croatian} (1) & \texttt{epitran} & \href{https://phoible.org/inventories/view/1139}{1139} & 305,112 & 1,109,696 & 39 \\
        Catalan & \href{https://childes.talkbank.org/access/Romance}{Romance/Catalan} (6) & \texttt{phonemizer} & \href{https://phoible.org/inventories/view/2555}{2555} & 319,726 & 1,084,594 & 36 \\
        Icelandic & \href{https://childes.talkbank.org/access/Scandinavian}{Scandinavian/Icelandic} (2) & \texttt{phonemizer} & \href{https://phoible.org/inventories/view/2568}{2568} & 279,939 & 1,057,235 & 35 \\
        Basque & \href{https://childes.talkbank.org/access/Other}{Other/Basque} (2) & \texttt{phonemizer} & \href{https://phoible.org/inventories/view/2161}{2161} & 230,500 & 942,725 & 49 \\
        Hungarian & \href{https://childes.talkbank.org/access/Other}{Other/Hungarian} (3) & \texttt{epitran} & \href{https://phoible.org/inventories/view/2191}{2191} & 237,062 & 918,002 & 48 \\
        Danish & \href{https://childes.talkbank.org/access/Scandinavian}{Scandinavian/Danish} (1) & \texttt{phonemizer} & \href{https://phoible.org/inventories/view/2265}{2265} & 275,170 & 824,314 & 42 \\
        Norwegian & \href{https://childes.talkbank.org/access/Scandinavian}{Scandinavian/Norwegian} (2) & \texttt{phonemizer} & \href{https://phoible.org/inventories/view/499}{499} & 227,856 & 729,649 & 43 \\
        PortugueseBr & \href{https://childes.talkbank.org/access/Romance}{Romance/Portuguese} (2) & \texttt{phonemizer} & \href{https://phoible.org/inventories/view/2207}{2207} & 174,845 & 577,865 & 44 \\
        Romanian & \href{https://childes.talkbank.org/access/Romance}{Romanian} (3) & \texttt{phonemizer} & \href{https://phoible.org/inventories/view/2443}{2443} & 152,465 & 537,669 & 43 \\
        Turkish & \href{https://childes.talkbank.org/access/Other}{Other/Turkish} (2) & \texttt{phonemizer} & \href{https://phoible.org/inventories/view/2217}{2217} & 79,404 & 421,129 & 51 \\
        Irish & \href{https://childes.talkbank.org/access/Celtic}{Celtic/Irish} (2) & \texttt{phonemizer} & \href{https://phoible.org/inventories/view/2521}{2521} & 105,867 & 338,425 & 34 \\
        Quechua & \href{https://childes.talkbank.org/access/Other}{Other/Quechua} (2) & \texttt{phonemizer} & \href{https://phoible.org/inventories/view/104}{104} & 46,848 & 281,478 & 40 \\
        Farsi & \href{https://childes.talkbank.org/access/Other}{Other/Farsi} (2) & \texttt{phonemizer} & \href{https://phoible.org/inventories/view/516}{516} & 43,432 & 178,523 & 40 \\
        \bottomrule
    \end{tabular}
    \caption{A breakdown of each language available in \ipachildes. The bracketed number in the \textbf{CHILDES Collection column} refers to the number of corpora downloaded from that collection. The \textbf{Backend}, \textbf{Lang Code} and \textbf{Phoneme Inventory} columns refer to the \gpp configuration used to convert utterances for that language to phonemes and the \phoible inventory used for that language in folding. The \textbf{Words} and \textbf{Phonemes} columns refer to the number of words and tokens in each subset and \textbf{\% Child} refers to the percentage of the data that is spoken by a child.}
    \label{tab:dataset-phonemized-childes-sections}
\end{table*}

\section{Dataset comparison}\label{app:datasetstats}

In \cref{sec:phonemicdatasets} we discuss previous phonemic datasets in relation to \ipachildes. We provide a full comparison of these datasets in \cref{tab:dataset-requirements}.

\setlength{\tabcolsep}{2pt}
\begin{table*}[t]
    \centering
    \small
    \begin{threeparttable}
        \begin{tabular}{lcccc}
             \toprule
            {\textbf{Dataset}} & {\textbf{Modality}} & {\textbf{Scale (words)}} & {\textbf{Domain}} & {\textbf{Languages}} \\
            \midrule
            The Pile \citep{pile} & Orth  & 100B\textsuperscript{\dagger}  & Web-scraped written text  & English only  \\
            GlobalPhone \citep{schultz2002globalphone} & Orth, Phon, Audio  & 5M\textsuperscript{\dagger}  & Read speech  & 22  \\
            CommonVoice \citep{ardila-etal-2020-common} & Orth, Audio  & 30M\textsuperscript{\dagger}  & Read speech  & 38  \\
            VoxCommunis \citep{ahn-chodroff-2022-voxcommunis} & Orth, Phon, Audio & 23M\textsuperscript{\dagger} & Read speech & 40 \\
            CMU Wilderness \citep{8683536} & Orth, Audio & 170M\textsuperscript{\dagger} & Read speech & 699 \\
            VoxClamantis \citep{salesky-etal-2020-corpus} & Orth, Audio, Phon & 152M\textsuperscript{\dagger} & Read speech & 635 \\
            TIMIT \citep{garofolo1993darpa} & Orth, Phon, Audio  & 40k & Read speech  & English only  \\
            FLEURS \citep{conneau2023fleurs} & Orth, Audio  & 15M\textsuperscript{\dagger}  &Read speech  & 102  \\
            MSWC \citep{mazumder2021multilingual} & Orth, Audio  & 20M  & Read speech  & 102  \\
            IPAPACK \citep{zhu-etal-2024-taste} & Orth, Phon  & 15M\textsuperscript{\dagger}  & Read speech  & 115  \\
            LibriSpeech \citep{panayotov2015librispeech} & Orth, Audio  & 10M\textsuperscript{\dagger}  & Audio books  & English only  \\
            Libri-Light \citep{Kahn_2020} & Orth,\textsuperscript{*} Phon,\textsuperscript{*} Audio  & 700M\textsuperscript{\dagger}  & Audio books  & English only  \\
            MLS \citep{pratap2020mls} & Orth,\textsuperscript{*} Phon,\textsuperscript{*} Audio  & 600M\textsuperscript{\dagger}  & Audio books  & 8  \\
            Switchboard \citep{godfrey1992switchboard} & Orth, Phon, Audio  & 3M\textsuperscript{\dagger}  & Telephone conversations  & English only  \\
            Fisher \citep{cieri2004fisher} & Orth, Audio  & 12M\textsuperscript{\dagger}  & Telephone conversations  & English only  \\
            Buckeye \citep{PITT200589} & Orth, Phon, Audio  & 300k  & Spontaneous speech  & English only  \\
            British National Corpus \citep{bnc2007} & Orth, Audio & 100M & Written \& spontaneous speech & English only \\
            Audio BNC \citep{coleman2012audio} & Orth, Phon, Audio  & 7M  & Spontaneous speech  & English only  \\
            VoxLingua107 \citep{9383459} & Audio & 80M & Spontaneous speech & 107 \\
            Babel \citep{harper2011babel} & Orth, Audio  & 60M  & Telephone conversations  & 25  \\
            CHILDES \citep{macwhinney1985child} & Orth  & 59M  & Child-centered speech & 45 \\
            BabyLM \citep{choshen-et-al-2024-callforpapers-babylm2} & Orth  & 100M  & Speech and text\textsuperscript{**} & English only  \\
            \midrule
            \ipachildes & Orth, Phon & 45M  & Child-centered speech & 31 \\
            \bottomrule
        \end{tabular}
        \normalsize
        \caption{A comparative summary of the datasets discussed in \cref{sec:phonemicdatasets}. The datasets are described in terms of their modality, scale, domain and languages. \ipachildes is the first multilingual phonemic dataset of spontaneous speech and the first phonemic dataset of child-centered speech. \\\emph{\textsuperscript{\dagger}Word counts estimated from the size in bytes or the hours of audio in the dataset, using a heuristic based on the size of Switchboard of 5 bytes per word and 12,000 words per hour.}\\\emph{\textsuperscript{*}Libri-Light and MLS only have orthographic and phonemic transcriptions for 10 hours of audio per language.}.\\\emph{\textsuperscript{**}BabyLM contains a mix of speech and text data from a mix of adult-directed and child-directed sources, only 29\% is child-directed speech.}}
        \label{tab:dataset-requirements}
    \end{threeparttable}
\end{table*}
\setlength{\tabcolsep}{6pt}

\section{\gpp Usage}\label{app:usage}

\gpp is a python library that can be used as an API or as a command-line tool in order to convert orthographic text to a phonemic representation. The tool allows the user to select the backend and language code to use for G2P with text provided through filepaths or standard input. Additional options include \verb|--keep_word_boundaries| to output a dedicated \texttt{WORD\_BOUNDARY} token between words and \verb|--uncorrected| to skip the folding process and output the phonemes exactly as produced by the backend tool. Each backend also supports individual options. For instance, \verb|--split-tones| outputs tones as individual tokens instead of merging them with the syllabic phoneme for our two Chinese language backends. See the repository's \texttt{README.txt} for further details.

\section{Phoneme Stream Representation}\label{sec:phonemestream}

In order to ensure that phonemes are output using a consistent representation, we define the \textbf{phoneme stream representation} as follows:
\begin{itemize}
    \item Each phoneme is represented using the International Phonetic Alphabet (IPA).
    \item Each phoneme is separated by a space.
    \item Word boundaries and utterance boundaries are represented using unique symbols.
\end{itemize}

IPA is used to represent each phoneme due to being the most widely used and comprehensive phonetic alphabet. It is important to separate phonemes by spaces because IPA symbols may be represented using multiple Unicode characters. For instance, the word ``enjoy'' can be transcribed in IPA as \textipa{EndZOI} which uses six characters but only contains four phonemes, since \textipa{dZ} is a single consonant and \textipa{OI} is a diphthong. By instead representing the word as \textipa{E~n~dZ~OI}, it is much easier to split the word into individual phonemes by using whitespace as a delimiter. Similarly, word boundaries and utterance boundaries are represented using the unique symbols \texttt{WORD\_BOUNDARY} and \texttt{UTT\_BOUNDARY}. 

\section{Folding Maps}\label{sec:folding-details}

Folding maps are primarily used to make surface-level adjustments, but they can also be used to solve several other error types in order to create a better alignment with a \phoible inventory. These errors are detailed in \cref{tab:transcription-errors}.

\begin{table*}[t]
    \centering
    \scriptsize
    \begin{tabular}{p{0.32\textwidth}p{0.26\textwidth}p{0.32\textwidth}}
    \toprule
        \textbf{Error type} & \textbf{Consequence} & \textbf{Example} \\
        \midrule
        \textbf{One-to-one:} The backend uses one symbol for a phoneme but the inventory lists a different symbol for that phoneme. & The one-to-one mapping does not change the number of types or tokens in the output. & \texttt{phonemizer} with language code \texttt{sv} (Swedish) outputs \ttipa{n} but the matching inventory uses \ttipa{\textsubbridge{n}}.\\
        \midrule
        \textbf{Many-to-one:} The backend produces two different phonemes that should only map to a single phoneme in the inventory. & The many-to-one mapping reduces the number of phoneme types. & \texttt{phonemizer} with language code \texttt{pt} (Portuguese) outputs both \ttipa{\*r} and \ttipa{r} but the matching inventory only lists \ttipa{K}.\\
        \midrule
        \textbf{Consonant merging:} The backend outputs two symbols for a consonant that should be written as a single phoneme. & The mapping merges the pair of consonants, reducing the number of phoneme tokens produced. & \texttt{epitran} with language code \texttt{srp-Latn} (Serbian) outputs the sequence \ttipa{d Z} but these are should be written as a single phoneme \ttipa{dZ}.\\
        \midrule
        \textbf{Vowel merging:} The backend outputs a pair of vowels as separate phonemes but they are typically analysed as a single diphthong. & The mapping merges the pair of vowels, reducing the number of phoneme tokens produced. & \texttt{pingyam} with language code \texttt{cantonese} outputs the sequence \ttipa{o u} but these are should be treated as a diphthong \ttipa{ou}.\\
        \midrule
        \textbf{Vowel splitting:} The backend outputs a diphthong that is not listed in the inventory and should be split into individual phonemes. & The mapping splits the pair of vowels, increasing the number of phoneme tokens produced. & \texttt{phonemizer} with language code \texttt{en-us} (North American English) outputs \ttipa{aIU} as a single phoneme but this should be \ttipa{aI U}.\\
        \midrule
        \textbf{Phoneme duplication:} The backend outputs duplicate phonemes to represent long vowels or consonants or because of an error. & The mapping replaces the pair of phonemes with just one, reducing the number of phoneme tokens. & \texttt{phonemizer} with language code \texttt{et} (Estonian) outputs \ttipa{d d} but should output the long consonant \ttipa{d:}.\\
        \midrule
        \textbf{Diacritic error:} The backend incorrectly outputs the diacritic as a separate symbol instead of attaching it to the phoneme. & The mapping may change the number of phoneme types or tokens. & \texttt{phonemizer} with language code \texttt{ko} (Korean) outputs the diacritic for aspiration as \ttipa{h} instead of \ttipa{\super{h}} so sequences \ttipa{kh} and \ttipa{ph} are mapped to \ttipa{k\super{h}} and \ttipa{p\super{h}}.\\
        \midrule
        \textbf{Orthographic error:} Due to an invalid symbol in the orthographic text, the backend outputs an incorrect phoneme. & The contextual mapping changes the frequency statistics for the resulting phoneme, possibly reducing the number of phoneme types. & \texttt{epitran} with language code \texttt{hun-Latn} (Hungarian) outputs \ttipa{\^o} when the orthographic letter \textipa{\H{o}} is incorrectly written as \textipa{\^o} and so the phoneme is mapped to \ttipa{\o:}.\\
        \bottomrule
    \end{tabular}
    \caption{A list of errors that can occur during grapheme-to-phoneme conversion that can be fixed with a folding map but that may change the information-theoretic properties of the output.}
    \label{tab:transcription-errors}
\end{table*}

The many-to-one mappings and those that split or merge tokens may alter the number of output tokens or types. Since such a mapping will change the information-theoretic properties of the output, it is important that they are linguistically motivated and carefully implemented. 

In order to construct the folding map for each backend-language pair, we run \gpp on orthographic text for that language and compare the output set of phonemes $P_O$ to the phonemes in the closest inventory in \phoible $P_I$. We call the set of phonemes present in $P_O$ but not $P_I$ the ``unknown phonemes'' $U_K$ where $U_K = P_O \setminus P_I $ and the set of phonemes present in $P_I$ but not $P_O$ the ``unseen phonemes'' $U_S$ where $U_S = P_I \setminus P_O $. We then construct the folding map as follows:
\begin{enumerate}
    \item Find pairs $(k,s) \in U_K \times U_S$ that differ according to an accent or diacritic and obviously represent the same phoneme (determined by ruling out alternatives or examining where $k$ is produced in the output). Create a one-to-one mapping $k:s$ for each such pair, e.g. \ttipa{t} : \ttipa{t\super{h}}.
    \item Find pairs $(k,s) \in U_K \times U_S$ that clearly represent the same phoneme (determined as above) but may use entirely different symbols, possibly due to an alternative transcription scheme. Create a one-to-one mapping for each pair, e.g. \ttipa{a} : \ttipa{\ae}.
    \item For remaining items $k \in U_K$, determine whether these result from one of the other errors in \cref{tab:transcription-errors}. Carefully examine instances where $k$ is produced in the output and create a suitable mapping $k : p$ for some $p \in P_I$ to solve the error (the mapping may need to be contextual or include several characters, e.g. \ttipa{\textrhookschwa} : \ttipa{@ \*r} or \ttipa{U O} : \ttipa{w O}). 
    \item For remaining items $s \in U_S$, determine whether these result from one of the other errors in \cref{tab:transcription-errors}. Carefully examine instances where $s$ should be produced in the output and create a suitable mapping $k : s$ for some $k \in P_O$ to solve the error (the mapping may need to be contextual or include several characters). 
    \item Examine the output for cases of \textbf{phoneme duplication} and other errors that may not contain phonemes in $U_K$ or $U_S$ but could still be solved with the phoneme map and create suitable mappings.
\end{enumerate}

The goal is for $U_K = \{\} = U_S$ or equivalently $P_I = P_O$, i.e the set of phonemes produced by the tool perfectly aligns with the phoneme inventory in \phoible. This is not always possible, often there are a few remaining phonemes in $U_K$ and/or $U_S$. This can occur when no obvious mappings could be found in steps 1--4 above. For example, the \texttt{epitran} backend for German does not produce the phoneme \ttipa{Z} (it is ``unseen'') and none of the unknown phonemes seem to be a good match. Another possibility is that the output set of phonemes $P_O$ may not align well with any of the \phoible phoneme inventories and so the closest match may not include some of the unknown phonemes $k \in U_K$ despite being valid phonemes for that language and listed in other inventories. For example, the \texttt{epitran} backend for German produce the phonemes \ttipa{x} and \ttipa{5} which are not listed in the matching inventory but are listed in other established inventories for German. In other cases, the unknown phonemes may come from loan words (e.g. \ttipa{ts} for ``pizza'' in Portuguese). Finally, there are some cases where the output considerably disagrees with all of the \phoible inventories but is a valid phonemic analysis of the language according to other sources.

See \cref{sec:qualitative} for an example of using \gpp for French, using the \texttt{phonemizer} backend with a folding map to approach Phoible inventory \href{https://phoible.org/inventories/view/2269}{2269}.

\section{Implementation Details}\label{sec:models}

We conduct our experiments using the \texttt{PyTorch} framework \citep{paszke-etal-2019-pytorch} and the \texttt{Transformers} library \citep{wolf-etal-2020-transformers}.

\subsection{Hardware Details}

We use a server with one NVIDIA A100 80GB PCIe GPU, 32 CPUs, and 32 GB of RAM for all experiments. Below, we report a subset of the output of the \emph{lscpu} command:

\begin{tcolorbox}[left=5pt,right=5pt,top=5pt,bottom=5pt]
\small
\begin{verbatim}
Architecture:        x86_64
CPU op-mode(s):      32-bit, 64-bit
Address sizes:       46 bits physical, 
                     48 bits virtual
Byte Order:          Little Endian
CPU(s):              32
On-line CPU(s) list: 0-31
Vendor ID:           GenuineIntel
Model name:          Intel(R) Xeon(R)
                     Silver 4210R CPU
                     @ 2.40GHz
CPU family:          6
Model:               85
Thread(s) per core:  1
Core(s) per socket:  1
Socket(s):           8
Stepping:            7
BogoMIPS:            4800.11
\end{verbatim}
\end{tcolorbox}

\subsection{Model Parameters and Training Procedure}

\begin{table}[h!]
    \centering
    \small
    \begin{tabular}{lcccc}
    \toprule
         Parameter & Value \\
         \midrule
         Max Example Length & 128 \\
         Learning Rate & 0.001\\
         Optimizer & AdamW \\
         Scheduler Type & Linear\\
         Max Steps & 200k \\
         Warm-up Steps & 60k \\
         Per Device Batch Size & 32 \\
    \bottomrule
    \end{tabular}
    \caption{Hyperparameter settings for training the GPT-2 architecture. Where values are not reported, they may be assumed to be default values.}
    \label{tab:training_params}
\end{table}

\begin{table}[h!]
    \centering
    \small
    \begin{tabular}{lccccc}
    \toprule
         Model Size & Layers & Heads & Embd & Inner \\
         \midrule
         400k & 2 & 4 & 128 & 512 \\ 
         600k & 3 & 4 & 128 & 512 \\ 
         800k & 4 & 4 & 128 & 512 \\ 
         1M & 6 & 4 & 128 & 512 \\ 
         5M & 6 & 8 & 256 & 1024 \\ 
         19M & 6 & 8 & 512 & 2048 \\ 
         25M & 8 & 8 & 512 & 2048 \\ 
         85M & 12 & 12 & 768 & 3072 \\ 
    \bottomrule
    \end{tabular}
    \caption{GPT-2 model sizes used in the size requirement experiment. Where values are not reported, they may be assumed to be default values.}
    \label{tab:model_sizes}
\end{table}

\begin{table*}[t!]
    \centering
    \small
    \begin{tabular}{c|ccc|ccc}
    \toprule
         Data Size& \multicolumn{3}{c|}{BabySLM Lexical} & \multicolumn{3}{c}{BabySLM Syntactic} \\
         (words) & Model Size & Dropout & Score & Model Size & Dropout & Score \\
         \midrule
         80k & 600k & 0.3 & 65.8 & 400k & 0.5 & 52.6 \\ 
         180k & 800k & 0.3 & 69.3 & 5M & 0.5 & 52.3\\ 
         500k & 5M & 0.3 & 72.9 & 5M & 0.3 & 54.3\\ 
         800k & 19M & 0.5 & 74.2 & 19M & 0.1 & 54.9 \\ 
         1.8M & 5M & 0.3 & 77.4 & 19M & 0.1 & 55.6 \\ 
         5M & 19M & 0.1 & 80.3 & 5M & 0.3 & 58.3 \\ 
    \bottomrule
    \end{tabular}
    \caption{Best model sizes and dropout values for the BabySLM Lexical and Syntactic scores for each subset size of the EnglishNA corpus of \ipachildes.}
    \label{tab:best_sizes}
\end{table*}

We describe training parameters in \cref{tab:training_params} and model sizes in \cref{tab:model_sizes}. Following the conventions of the Pythia suite of models \citep{biderman2023pythia}, we report the number of non-embedding parameters. Unlike their suite, where models are named according to the number of parameters, we name our models according to the number of non-embedding parameters. This is because we use the same architecture for multiple languages, each of which has a different vocabulary size according to the number of phoneme types in that language, which alters the total number of parameters. Our 1M, 19M and 85M models are equivalent to Pythia-14M, Pythia-70M and Pythia-160M, respectively. Our training scripts are available \href{https://github.com/codebyzeb/PhonemeTransformers}{here}.

Data is prepared into batches by first tokenizing the entire dataset, combining all tokens into one long vector, and then splitting the vector into chunks of 128 tokens. Only the very last example is padded, if required. At each step during training, random chunks are selected and combined into batches. 

Checkpoints are taken every 20,000 steps during training. At each checkpoint, the perplexity is evaluated on the held-back evaluation set, and at the end of training the checkpoint with the lowest perplexity is returned as the best model. For the smallest models, many of the best models were from the very first checkpoint, since due to the small training dataset and small model, the model had already fit the data by this point.

In our size requirement experiment (see \cref{sec:sizerequirements}), we train each model in \cref{tab:model_sizes} using a dropout of 0.1, 0.3 and 0.5 on each subset size of the EnglishNA portion of \ipachildes.

\section{Best Phoneme LM Parameters Across Data Scales}\label{sec:best-model-parameters}

Following the size experiment in \cref{sec:sizerequirements}, we report the model size and dropout values that achieved the highest BabySLM scores for each subsample size of the EnglishNA portion of \ipachildes in \cref{tab:best_sizes}. 

\section{Average Information Density of Phonemized Child-Directed Speech Increases with Age Cross-Lingually}\label{app:parentese}

The phonemic representation of the utterances in our dataset open up new avenues for exploring the phonotactic properties of languages and the information-theoretic properties of child-directed speech. 
\begin{figure}
    \centering
    \includegraphics[width=0.99\linewidth]{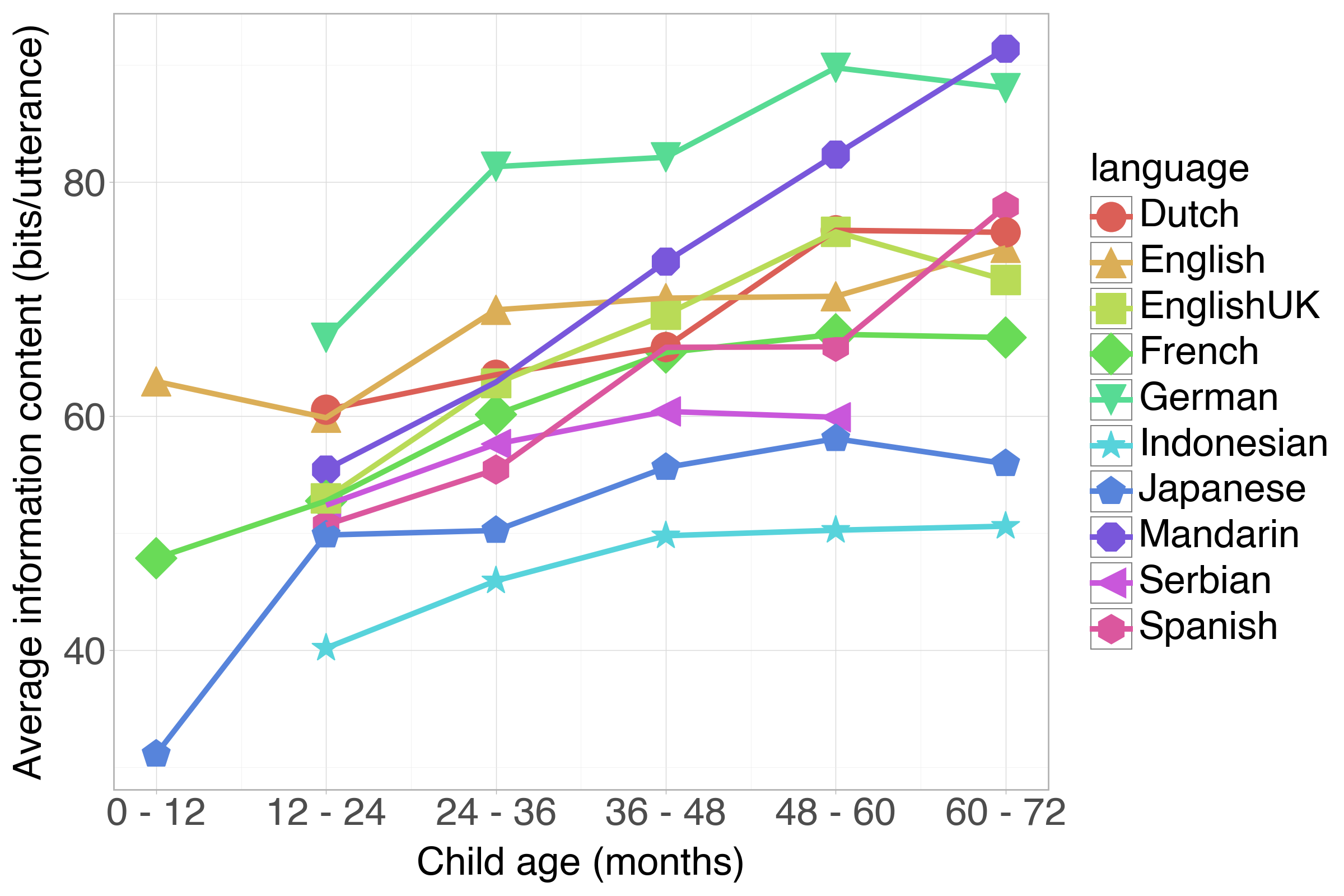}
    \caption{Average information of child-directed utterances in CHILDES}
    \label{fig:information-trends}
\end{figure}

Here, we demonstrate one information-theoretic experiment, comparing the average information content of child-directed utterances to the age of the child being spoken to (this information is also available in CHILDES and is preserved in our dataset). We group child ages in years (0-12 months, 12-24 months, etc.) and calculate the average information content of a sample of child-directed utterances using a unigram language model. The information $I_U$ of each utterance consisting of a sequence of phonemes $p_1,p_2,\ldots,p_n$ is given by

$$I_U = -\sum_{i=0}^{n}{log_2P(p_i)},$$

where $P(p_i)$ is the probability of phoneme $p_i$ given by its frequency in the data. We plot the average information of utterances in each age category for the largest 10 languages in the dataset in \cref{fig:information-trends}. We find that across all 10 languages the average information of utterances increases with the age of the child, indicating that speakers of `Parentese' may adjust the complexity of their speech according to the learner's age.

\end{document}